\documentclass[10pt,twocolumn,letterpaper]{article}

\usepackage{iccv}
\usepackage{times}
\usepackage{epsfig}
\usepackage{graphicx}
\usepackage{amsmath}
\usepackage{amssymb}
\usepackage[accsupp]{axessibility} 
\usepackage{subfigure}

\usepackage{algorithm}
\usepackage{algorithmic}
\usepackage{array}
\usepackage{pifont}
\usepackage{wrapfig}
\usepackage{multirow}
\usepackage{setspace}

\usepackage{booktabs}
\usepackage{mathtools}

\usepackage[pagebackref=true,breaklinks=true,letterpaper=true,colorlinks,bookmarks=false]{hyperref}

\iccvfinalcopy % *** Uncomment this line for the final submission

\def\iccvPaperID{7253} % *** Enter the ICCV Paper ID here

\newcommand{\dec}[0]{\texttt{dec}}
\newcommand{\cmark}{\ding{51}}%
\begin{document}

%%%%%%%%% TITLE
\title{Probing Visual-Audio Representation for Video Highlight Detection via Hard-Pairs Guided Contrastive Learning}

\author{Shuaicheng Li$^{1*}$, Feng Zhang$^{1*}$, Kunlin Yang$^{1}$, Lingbo Liu$^{2}$, Shinan Liu$^{1}$, Jun Hou$^{1}$, Shuai Yi$^{1}$\\
$^{1}$Sensetime Research, $^2$The Hong Kong Polytechnic University, China
\\
{\tt\small \{lishuaicheng,zhangfeng4,yangkunlin,liushinan,houjun,yishuai\}@sensetime.com}\\
{\tt \small liulingbo918@gmail.com}
%\and
}

\maketitle
\ificcvfinal\thispagestyle{empty}\fi
\begin{abstract}
  Video highlight detection is a crucial yet challenging problem which aims to identify the interesting moments in untrimmed videos.
  %Effectively addressing the problem of video highlight detection necessitates cross-modality representations that jointly pursues two goals, \textit{i.e.}, cross-modal representation learning and fine-grained feature discrimination.
  The key to this task lies in effective video representations that jointly pursue two goals, \textit{i.e.}, cross-modal representation learning and fine-grained feature discrimination.
  In this paper, these two challenges are tackled by not only enriching intra-modality and cross-modality relations for representation modeling but also shaping the features in a discriminative manner.
  Our proposed method mainly leverages the intra-modality encoding and cross-modality co-occurrence encoding for fully representation modeling.
  Specifically, intra-modality encoding augments the modality-wise features and dampens irrelevant modality via within-modality relation learning in both audio and visual signals.
  Meanwhile, cross-modality co-occurrence encoding focuses on the co-occurrence inter-modality relations and selectively captures effective information among multi-modality.
  The multi-modal representation is further enhanced by the global information abstracted from local context.
  In addition, we enlarge the discriminative power of feature embedding with a hard-pairs guided contrastive learning (HPCL) scheme.
  %, which enforces the similar sample closer and push away dissimilar ones for shaping structure-aware embedding space.
  %In addition, we propose a dense segment-wise contrastive learning scheme for enlarging the discriminative power of feature embedding under the supervised setting.
  %It enforces the similar sample closer and push away dissimilar ones for shaping structure-aware embedding space.
  A hard-pairs sampling strategy is further employed to mine the hard samples for improving feature discrimination in HPCL.
  Extensive experiments conducted on two benchmarks demonstrate the effectiveness and superiority of our proposed methods compared to other state-of-the-art methods.
  %outperforms other state-of-the-art methods and demonstrate the effectiveness of our proposed methods.
  %\textit{We will release our source code in the final version.}
\end{abstract}

\section{Introduction}
\let\thefootnote\relax\footnotetext{$*$ \textrm{\ indicates\ equal\ contribution.}}

\begin{figure}[t]
\centering
\includegraphics[width=1.0\linewidth]{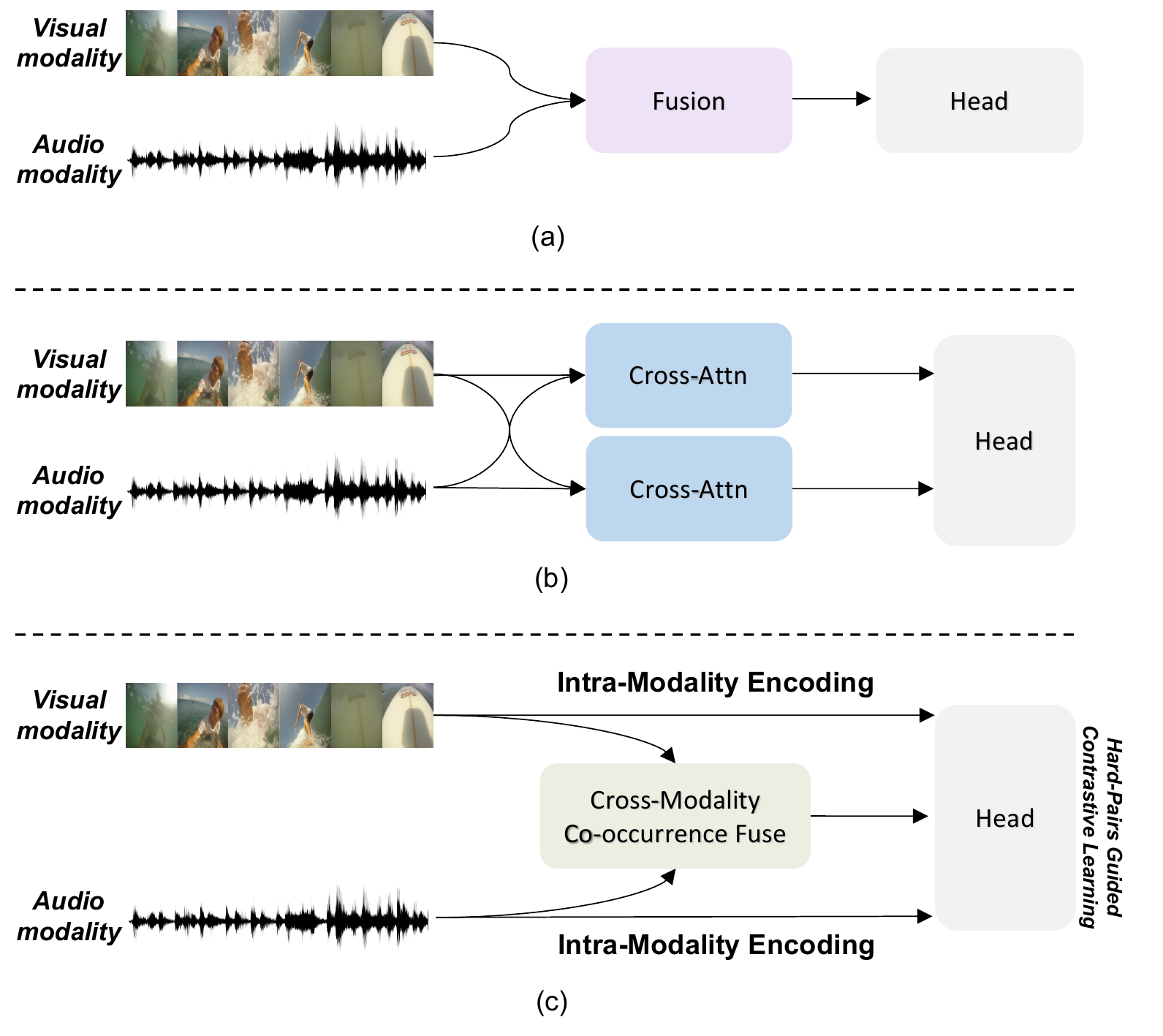}
\caption{{Schematic depiction of multi-modal representation learning. (a) simple fusion scheme. (b) within-modality learning with cross-attention. (c) Our proposed method explores the inter and intra-relations by measuring intra-visual, intra-audio and cross-modality co-occurrence. And HPCL is introduced for feature discriminative.}
}
\label{fig:intro}
\end{figure}
In recent years, posting well-edited video shining moments on social platforms, \eg, Youtube, Tiktok, has become our daily routine. 
Due to the labor for cropping untrimmed videos, video highlight detection task has drawn extensive attention from the research community.
The goal of this task is to localize the highlight segments and trim shining moments from untrimmed long videos automatically, which has a wide range of downstream applications such as video summarization \cite{hong:2020:mini=net,LSVM} and detecting \cite{tirupattur:2021:modeling-TAL,li2021groupformer}.

To identify the interesting part of an untrimmed video sequence, approaches make sorts of efforts to discriminate highlight and non-highlight clips.
%They are mainly divided into two categories.
%The majority of existing methods are committed to exploring visual highlight detection.
Pair-based approaches \cite{yao:2016:rankfirst,sun:2014:ranking,kim:2018:exploitingrank,jiao:2018:three} assumed that there exists distinguishable appearances between highlight and background segments. 
A ranking model was trained based on pairs (\textit{highlight non-highlight}) to rank segment scores and select shining moments. 
\cite{zhang:2020:find,xu:2021:crossSL,chen:2021:pr} captured contextual information to predict segment-level scores and generate highlights.
Nevertheless, these methods only utilized visual appearance and hindered the ability of modeling useful video representation among video segments.
Thus, \cite{tian:2020:unified,badamdorj:2021:joint,hong:2020:mini=net} developed an audio-visual network to assemble multi-modal representations, indicating that better representation modeling benefits more for highlight detection performance.
%Basically, these models usually project video segments to non-linear embedding space and then generate a binary label for each segments to predict whether it is highlight moment, indicating that better representation modeling benefits more for highlight detection performance.
Therefore, an essential problem can be asked: \textit{How to fully exploit video representations?}

Intuitively, it should not only (1) capture multi-modality contextual information, but also (2) be well distinguishable to inter-segments.
\textbf{To issue (1)}, a main stream of efforts delves into effective feature learning, \eg, cross-modal signals fusion \cite{badamdorj:2021:joint,hong:2020:mini=net,noda2015audiocc}.
There are two directions on handing multi-modal data.
The first is to extract visual and audio features and aggregate them simply as shown in Figure \ref{fig:intro} (a).
The second is modeling cross-modality representations by cross-attention modules (Figure \ref{fig:intro} (b)) such as \cite{badamdorj:2021:joint,tian:2020:unified,chen:2020:uniter}.
Nevertheless, these methods are sub-optimal for exploiting the complex relationships between inter-modality since these methods are based on the assumption that multiple signals are synchronized, which may not hold in practice with spurious noise and indistinct correspondence between these modalities.
%there exists spurious noise and indistinct correspondence between these signals.
\textbf{With regard to issue (2)}, prior studies \cite{sun:2014:ranking,yao:2016:rankfirst,kim:2018:exploitingrank} employ ranking models to facilitate segment pairs discrimination.
They only push away the dissimilar pairs by ranking loss and do not reflect intrinsic semantic representation.
Moreover, since there exists highly similar content for consistent video segments, it is essential to focus on the distinction between highlight segments and its surrounding non-highlight clips.
%So we conclude that it can be a better strategy to address feature discrimination.

To address the above challenges, this paper goes deeper into designing visual-audio architectures by two views: (1) \textit{cross-modal representation learning} and (2) \textit{inter-segment feature discrimination}.
We propose a novel visual-audio framework for highlight detection.
%Attention-based manner eliminates the temporal inductive bias and enables long-term contextual information learning.
%solo audio stream
%triple
It consists of audio, visual and co-occurrence stream to not only explore their inter and intra-relationships among segments but also exclude unrelated clues.
%In a nutshell, audio and visual stream process appearance and audio signal separately by an encoder-decoder manner. 
%For audio and visual stream, we first map the inputs to a set of segment-level vectors by an original transformer encoder
By capturing the relationships from intra-visual, intra-audio and co-occurrence cross-modality with modal-wise and sequence-wise attention module, we can fully exploit segment-level embedding, and the correlated highlight segments can be correctly determined. 
We further introduce a decoder to abstract global context by integrating the segment-level representations.
In the latter view (2), we propose a hard-pairs guided contrastive learning (HPCL) scheme for segment-to-segment embedding discrimination instead of image-wise training strategy.
HPCL treats the video highlight detection task as a dense segment-level embedding classification, aiming to improve intra-class compactness and inter-class dispersion.
In detail, HPCL shapes segment embedding space in a discriminative manner by pulling in similar samples against dissimilar ones.
Under the supervised setting, the categorical information is given during training, thus the positive data is the highlight segments while the negative ones are the non-highlights. 
We view every segment as anchor, and apply HPCL to learn the structural representations of labeled segments.
Furthermore, for a video sequence, the most indistinguishable segments always emerge from highlight boundaries due to consecutive appearance content.
Therefore, a hard-pairs sampling strategy is introduced to improve the discriminating power by mining these negative segments in our HPCL.
%To demonstrate the effectiveness of our proposed method, we evaluate the proposed method on two widely used benchmarks (\ie, \textit{Youtube Highlights and TVsum}) and achieve superior detection results compared to previous methods.
Finally, we conduct extensive experiments on two widely used benchmarks (\ie, \textit{Youtube Highlights and TVsum}), and the results show that our method achieves superior detection performance compared to previous methods.
The main contributions can be summarized as follows:
\begin{itemize}
    \item We propose a novel visual-audio framework to capture intra-visual, intra-audio and co-occurrence cross-modality information using modal-wise and sequence-wise attentions. 
    The global contextual representation is further exploited to enhance the video features.
    \item We develop a hard-pairs guided contrastive learning scheme in the fully supervised setting for video highlight detection.
    It shapes the video embedding space in a discriminative manner to reflect structural representation of video sequence.
    Besides, a hard-pairs sampling strategy is introduced to improve the discriminating power in HPCL by mining confused segments.
    \item Extensive experiments are conducted on the YouTube Highlights and TVsum benchmarks, and our proposed method outperforms other state-of-the-art methods.
    Detailed ablation studies demonstrate the effectiveness of our novel components.

\end{itemize}

\section{Related Works}
\subsection{Video Highlight Detection.}
The goal of the video highlight detection task is to predict the highlight moments according to the semantic features on the untrimmed videos. 
Recently, video highlight detection achieves extensive attention and is widely used in various types of videos, \textit{i.e.} sports video\cite{sprot-video}, first-person video\cite{yao:2016:rankfirst}. 
Mainstream methods \cite{sun:2014:ranking,yao:2016:rankfirst,kim:2018:exploitingrank} treat the video highlight detection task as a pair-based ranking.
They make the prediction scores of the highlight segment higher than those of the non-highlight segments by optimizing ranking loss.
%However they isolate individual segments from each other and does not take advantage of the rich contextual information in the untrimmed videos. 
Recent methods\cite{badamdorj:2021:joint,xu:2021:crossSL} propose to use self-attention mechanism to capture contextual features.
These methods utilize the temporal relations between segments and achieve excellent performance.
%Nevertheless, since massive annotations may be a time-consuming and laborious work, weakly supervised highlight detection utilize video-level labels and effectively reduce the demand for annotation labors.
\cite{xiong:2019:lessmore} finds that a clip in a shorter video is more likely to be a highlight clip, and then develops a ranking network with this prior knowledge. 
The method \cite{hong:2020:mini=net} treats highlight detection as multi-instance learning and introduces max-max ranking loss to better model video-level features and achieves high-quality results. 
Furthermore, MINI-Net\cite{hong:2020:mini=net} utilizes the visual-audio feature fusion to augment the video information by simple concatenation of each clip.
Joint-VA \cite{badamdorj:2021:joint} develops a cross-attention module following other works \cite{tian:2020:unified,sterpu:2018:attention-cross} to exploit cross-modal features and then utilizes noise sentinel to relieve the feature confusion from multi-modality.
%Due to the effectiveness of multi-modal fusion, %\cite{badamdorj:2021:joint} exploits \textit{how to allow each modality to modulate the other?}
%A cross attention module is applied to exploit cross-modal features and the noise sentinel is used to relieve the feature confusion from multi-modality.
Different from these approaches, our methods propose to explore intra- and inter-modality co-occurrence information and global context for semantic features learning.
On the other hand, we introduce a dense contrastive learning scheme to improve intra-class compactness and inter-class dispersion for better feature modeling, which is the first one to solve the feature discrimination in this task.
%%TODO ours 

%By studying the interrelationships between representations of different modalities, some methods utilize information from multiple modalities to achieve more accurate highlight detection models. 
%[?] use simple feature concatenation to achieve feature fusion of two modalities. 
%Compared with this crude fusion method, [?] optimizes the fusion process.
%They use the cross-attention mechanism[?] to allow interaction between different modalities, and use an adaptive weighting  summation method to complete the fusion of multiple modal features.

\subsection{Video Summarization}
The purpose of video summarization is to generate compact visual summaries from a complete video so that the amount of information in the video can be reflected to the greatest extent.
Thus video summarization is highly relevant to video highlight detection for finding shining momments.
Due to the time-consuming annotations, early methods \cite{lu:2013:sum-story,kim:2014:sum-reconstructing,lee:2012:sum-discovering,khosla:2013:sum-large} usually focus on unsupervised learning.
They mainly rely on hand-crafted heuristics for feature learning  and then utilize the segment-wise scores to select representative frames.
Thanks to user-labeled videos on the web and the development of deep learning, supervised deep learning methods have developed rapidly in recent years.
For instance, vsLSTM \cite{vslstm} regards the video summarization as sequential data learning and models the variable-range relations using two LSTMs \cite{hochreiter1997:lstm}. 
In order to reduce the impact of insufficient data annotation, some weakly supervised video summarization methods utilize video-level annotation information. 
\cite{mahasseni:2017:sum-unsupervised} proposes a generative adversarial network, which consists of summarizer and discriminator, to reconstruct video appearance and then learns to distinguish the input videos and its reconstructions.
Since the highlight detection is closely related to video summarization task, following the previous methods \cite{hong:2020:mini=net,badamdorj:2021:joint,xu:2021:crossSL}, we also utilize the widely adopted benchmarks in video summarization tasks, \ie, TVSum \cite{tvsum} to evaluate the effectiveness of our model.

\subsection{Multi-modal Representation}
Recently, multi-modal representation learning has been rapidly developed. 
A variety of methods\cite{lu:2019:vilbert,li:2019:visualbert,li:2022:clip-event,radford:2021:clip} adopt two-stream pre-training architectures, which contain two separate transformer encoders, to encode multi-modality relations for a few of downstream tasks such as action recognition \cite{gao2020listen}, speech recognition\cite{afouras:2018:audio-deep,sterpu:2018:attention-cross} and salient object detection \cite{zhang:2019:memory-objectidetection,zhang:2020l:fnet-objectidetection}.
%Nevertheless, separately capture modal-aware representation may be sub-optimal.
Correlations between modalities provide rich information for representation learning.
Cross-attention module\cite{li:2019:visualbert,badamdorj:2021:joint} is usually employed to capture semantic correlations.
And TCG\cite{ye:2021:TCG} develops a low-rank audio-visual tensor fusion mechanism to capture the complex association between two modalities, which can generate informative audio-visual fused features.
These works are usually based on the assumption that audio and visual data are synchronized and highly correlated \cite{owens2018:audio-sy,korbar2018:cooperative-sy}.
It may not hold in practice with indistinct correspondence between inter-modality.
We utilize the segment-wise attention to selectively capture the fine-grained relations between inter-modality and dampen the noise in both modalities.

%\textit{i.e.} using text information to assist object detection [?], using audio input to enhance action recognition [?], and the current popular multi-modal pre-train task[?]. 
%Research in [43] shows that using simple multi-modal feature fusion methods often leads to suboptimal results.
%Therefore, some researchers focus on methods to improve feature fusion [?]. 
%However, existing methods still cannot effectively deal with the problem of domain misalignment between multimodal features, and our proposed method effectively alleviates this phenomenon.

\subsection{Contrastive Learning}
Many studies \cite{van:2018:contrast-rep,he:2020:moco,chen:2020:simplecontrats,chen:2020:mocov2} on unsupervised representation learning concentrate on the central concept: contrastive learning.
%DenseCL \cite{wang2021denseCL} applies the contrastive objective to the pixel-level pre-training.
\cite{zhang2020selfcontrast} develops hierarchical segmentation algorithm for grouping pixels.
\cite{van2021unproposal} leverages object masks from pre-trained saliency detectors to group samples.
%. Recently, some works employed contrastive learning for unsupervised semantic segmentation by grouping the pixels
%with the help of hierarchical segmentation algorithm [61]
%or off-the-shelf saliency module [47]
%Contrastive methods learn representations in a discriminative manner by pushing away dissimilar (negative) pairs while pulling in similar (positive) data pairs in unsupervised setting.
They generate several positive augmented version by perturbations while negative data are randomly sampled from the other images.
%In this paper, 
%we extend the standard contrastive learning to a dense segment-wise paradigm (HPCL) in supervised learning setting instead of the above unsupervised pre-training method.
The above-mentioned works typically consider contrastive learning as pre-training step and use the variant versions as positive samples in unsupervised setting,
Different from these methods, we raise a segment-wise dense contrastive learning scheme in the fully supervised setting with the known categorical information for highlight detection.
Moreover, we also present a hard-pairs sampling strategy tailored for our video task to enlarge the discriminative power of our method.

\begin{figure*}[t]
%\begin{wrapfigure}{r}{0.5\textwidth}
  \centering
  \includegraphics[width=1.0\linewidth]{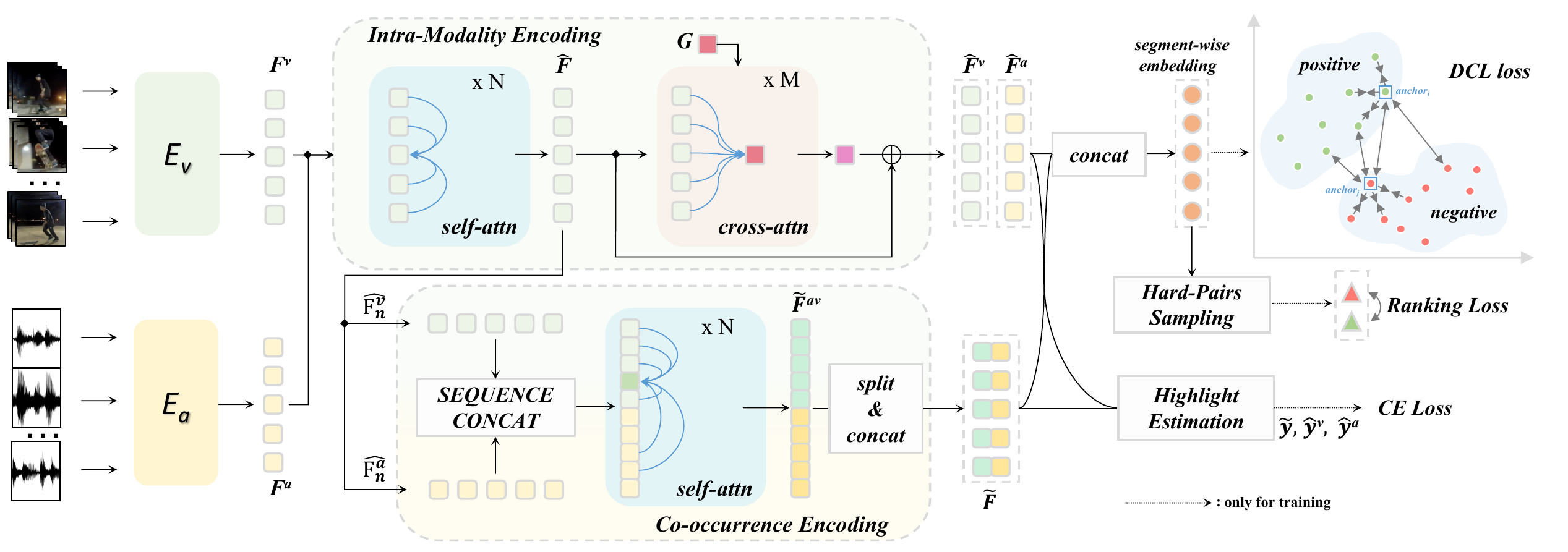}
  \caption{
  Illustration of our proposed method.
  $E_v$ and $E_a$ extract the high-level visual and audio features respectively and then followed the intra-modality encoding for modeling visual and audio representation separately.In addition, the cross-modality co-occurrence encoding is employed to exploit inter-modality relations.
  Lastly, for the output embeddings, we view segment-level representation as a point, a dense contrastive learning is proposed to shape the structural information in a discriminative manner.
  }
  \label{fig:arch}
\end{figure*}

\section{Methods}
%Before detailing our dense contrasive learning scheme (Sec. \ref{sec:3-7}), we first introduce the Intra-Modality Encoding (Sec. \ref{sec:3-4}) and Co-Occurrence Encoding (Sec. \ref{sec:3-5}) for cross-modality encoding.
%We start by summarizing the overall architecture in Sec. \ref{sec:problem-definition}, and then elaborate on the basic modules in Sec. \ref{}
%\subsection{Problem definition}
%\label{sec:problem-definition}
Given an arbitrary unedited video sequence $V=\{ v_t \}_{t=1}^{T}$ containing $T$ segments, each segment $v_t$ is annotated as binary label $y_t \in \{ 0, 1 \}$, indicating whether $v_t$ contains the interesting part about categorical moments.
Models are aiming to predict the label (\textit{\ie, highlight or non-highlight}) of every segment.
%\subsection{Overview}
Our proposed method is illustrated in Figure \ref{fig:arch}.
Firstly, we extract high-level features for evert segment using feature extractor in Sec. \ref{sec:feat-extract}.
the CNN backbones $E_v$ and $E_a$ are utilized to extract visual and audio features respectively, which can be represented as 
visual features $\mathbf{F}^v=\{ f_1^{v},..., f_T^v\} \in \mathbb{R}^{T \times d}$ and audio features $\mathbf{F}^a=\{ f_1^{a},..., f_T^a\} \in \mathbb{R}^{T \times d}$.
$d$ is the feature dimension. 
Secondly, the visual and audio features are fed into separate Intra-Modality Encoding to contextualize informative features.
In detail, we use self-attention modules to capture within-modality features, namely $\mathbf{F}^v {'}$, $\mathbf{F}^a {'}$, and then update the embeddings by abstracting the global context.
After that, to pursue distinct associations in inter-modality, a cross-modality Co-occurrence Encoding is developed to capture the fine-grained relations for representation learning.
Finally, the output representations of Co-occurrence Encoding, together with the intra-modality features are used to generate the segment-level confidence scores by three classifiers respectively and obtain the final highlight detection results by the weighted sum of final scores.
%With these multi-modal features, our visual-audio network respectively enhances these multi-modal representations by intra-visual, intra-audio and cross-modality co-occurrence relations.
%We also contextualize the global features for updating the segment representations.
%A straightforward way to enrich segment representation is to concatenate the global features and segment representations.
%Finally, the refined segment-level representations are used to generate segment-level confidence scores by a classifier and obtain the final highlight detection results.
%TODO  not write HPCL
On the other hand, we introduce the dense contrastive learning scheme to shape structural representations in the training phase.
Specifically, for the output segment-wise representations of Intra-Modality Encoding, we utilize dense contrastive loss to compute segment-to-segment contrast to regularize the latent embedding space and use the hard-pairs sampling strategy to improve the discrimination power.

\subsection{Feature Extractor}
\label{sec:feat-extract}
Given an untrimmed video sequence, the visual features are extracted by a 3D CNN \cite{hara:2018:3dcnn} with ResNet-34 \cite{he:2016:res} backbone $E_v$ pretrained on the Kinetics-400 dataset \cite{carreira:2017:kinetics}, while audio information is extracted by PANN \cite{kong2020:panns} audio network $E_a$ pretrained on AudioSet \cite{gemmeke2017audio} following the previous methods \cite{badamdorj:2021:joint,wang:2020:full-length}.
The visual and audio features of each segment are then flattened into a feature vector and are further transformed to the same embedding space with a linear layer respectively. 
Thus, the visual and audio features of the whole video can be denoted as $\mathbf{F}^v \in \mathbb{R}^{T \times d}$ and $\mathbf{F}^a \in \mathbb{R}^{T \times d}$ respectively, where the feature dimension d is set to 256 in this work.
%Visual and audio features of each segment are flattened into a feature vectors respectively.
%And then, the visual and audio features are linearly transformed to the same embedding space, which denote as $\mathbf{F}^v \in \mathbb{R}^{T \times d}$ and $\mathbf{F}^a \in \mathbb{R}^{T \times d}$ respectively.
%Here, we set $d=256$.
Moreover, instead of 3D CNN, we also employ the ImageNet \cite{krizhevsky:2012:imagenet} pretrained I3D \cite{Carreira:2017:I3D} backbone to extract visual segment features and conduct experiments.
It achieves better results than 3D CNN\cite{hara:2018:3dcnn} in practice due to the large perceptive field.

\subsection{Intra-Modality Encoding}
\label{sec:3-4}

We leverage the standard transformer encoder \cite{attention2017} to model within-modality relations and dampen the irrelevant modality.
Our Intra-Modality Encoding is designed by stacked $n$ encoder layers.
The modality-wise encoders are deployed to embed contextual features.
For visual stream, the input embeddings of $i$-th encoder is represented as $\textbf{F}^v_{i-1}$.
The first encoder takes the output features $\mathbf{F}^v=\{ {f}_1^{v},..., f_T^v\} \in \mathbb{R}^{T \times d}$ from feature extractor as input, which denotes as  $\textbf{F}^v_{0}=\textbf{F}^v$.
%For the visual input $\mathbf{F}^v=\{ {f}_1^{v},..., f_T^v\} \in \mathbb{R}^{T \times d}$, we first transform the input to three parts, \textit{i.e.}, queries $Q^v \in \mathbb{R}^{T \times d_k}$ , keys $K^v \in \mathbb{R}^{T \times d_k}$ and values $S^v \in \mathbb{R}^{T \times d_v}$.
In detail, in terms of the $i$-th encoder layer, we first transform the input to three parts, \textit{i.e.}, queries $Q^v \in \mathbb{R}^{T \times d_k}$ , keys $K^v \in \mathbb{R}^{T \times d_k}$ and values $S^v \in \mathbb{R}^{T \times d_v}$.
Here $d_k$, $d_v$ denote the dimensions of queries(keys) and values separately.
And then, we apply a multi-head dot-product attention block to capture intra-modality segment-wise relations and a feed-forward network (FFN) for fearure transformation and non-linearity.
This process can be formulated as,
\begin{align}
    &Q^v=W^q_i \! \mathbf{F}^v_{i-1},\quad \! K^v=W^k_i \mathbf{F}^v_{i-1},\quad \! S^v=W^s_i \mathbf{F}^v_{i-1} \\
    &\mathbf{F}^v_{i-1}=\mathrm{softmax}(\frac{Q^v {K^v}^{\mathrm{T}}}{\sqrt{d_k}})S^v + S^v\\
    %\phi_{attn}(Q,K,S)=\mathrm{softmax}(\frac{Q^v {K^v}^{\mathrm{T}}}{\sqrt{d_k}})S^v \\
    &\mathbf{F}^v_{i}=\mathrm{FFN}(\mathbf{F}^v_{i-1}{'})
\end{align}
where $\mathbf{F}^v_{i}$ is the output of the $i$-th modality encoder layer, $W_i^q, W_i^k \in  \mathbb{R}^{d_k \times d}$ and $W_i^s \in \mathbb{R}^{d_v \times d}$ are learnable parameters.
%$W_1,W_2 \in \mathbb{R}^{d \times d}$ and $b_1, b_2 \in \mathbb{R}^{d}$ are learnable parameters.

%Therefore, the output of n-th encoding layer $F^v_n$ is given by
%\begin{align}
%    \mathbf{F}^v'_n=\phi_{attn}&(W^q_n \mathbf{F}^v_{n-1}, W^k_n \mathbf{F}^v_{n-1},W^s_n \mathbf{F}^v_{n-1}), \\
%    &\mathbf{F}^v'_n=FFN(\mathbf{F}^v'_n),
%\end{align}
%where $W^q_n, W^k_n, W^s_n \in \mathbb{R}^{d \times d}$ are learnable parameters and $\mathbf{F}^v_{n-1}$ is the visual feature output from the previous attention encoding layer.

Despite the above encoders are capable of modeling intra-modality information, abstracting the global context is critical to refine video information.
Previous weakly supervised method \cite{hong:2020:mini=net} distributes the video-level features to every segment (\textit{global to local fashion}), which demonstrate its effectiveness of video-level information.
Thus, we speculate that summarizing segment features as global contextual information benefits feature embedding.
We introduce a decoder to parse the visual features $\mathbf{F}^v_n$ (omit the layer number $n$ for clarity) and contextualize the global features where the decoder is implemented with the pure transformer decoder containing multi-head cross-attention blocks and feed-forward network.
Motivated by the learned query proposed by \cite{DETR2020}, we formulate a learnable parameters $\mathbf{G}_\texttt{init} \in \mathbb{R}^d$ as initial global context.
%Then, we utilize the decoder to aggregate the video representations to $\mathbf{G}^v$.
The decoder takes the updated visual features $\mathbf{F}^v_n$ and $\mathbf{G}_\texttt{init}$ as input.
It views $\mathbf{G}_\texttt{init}$ as \textit{query} and visual features $\mathbf{F}^v_n$ as \textit{values}, and then the \textit{query} aggregates the visual context information and abstracts global informative representation represented as $\mathbf{G}^v$.
Finally, a straightforward method is to directly sum the global context and video representations, which can be formulated as $\mathbf{\hat{F}}^v=\mathbf{F}^v_n + \mathbf{G}^v$.

In the audio stream, the other intra-modality encoding module is also applied to refine the input audio features $\mathbf{F}^a$, and then generate the refined audio representations $\mathbf{\hat{F}}^a=\mathbf{F}^a_n + \mathbf{G}^a$ following the implementation of visual stream.

%%TODO add more contents

\subsection{Co-occurrence Encoding}
\label{sec:3-5}
Previous works \cite{chen:2020:uniter,hong:2020:mini=net,badamdorj:2021:joint} have confirmed that exploiting cross-modal relationships can make the representations better, and they usually utilize cross-modal encoder to capture semantic associations based on these multi-modal signals.
However, it can be sub-optimal since these works are usually based on the assumption that multi-modal signals are synchronized which may not hold in practice due to indistinct correspondence between these modality.
Audio or visual shining events can be partially presented at a single segment \cite{tirupattur:2021:modeling-TAL}.
Ideally, we would like the proposed method can dampen the noise and selectively choose effective information from multi-modality instead of all in them.
Therefore, we employ the sequence-wise attention module to capture the fine-grained relations among the visual and audio embeddings.
It can relieve the inter-modality asynchronization by learning to ignore the cross-modal segment features with spurious noise and augment the intimate ones.
Our cross-modality co-occurrence encoding is also built upon the attention mechanism in canonical transformer decoder and takes the full sequence of segment embeddings corresponding to all visual and audio features $\mathbf{F}^v_n =\{f^v_1{'}, ..., f_T^v{'}.\}, \mathbf{F}^a_n=\{f^a_1{'}, ..., f_T^a{'}.\}$ as input.
Assume the sequence modalities input $\mathbf{F}^{va}=\{f^v_1{'}, ..., f_T^v{'}, f_1^a{'}, ..., f_T^a{'} \} \in \mathbb{R}^{2T \times d}$, we alternatively contextualize the co-occurrence information for each modality.
In the visual modality, the process can be defined as,
\begin{align}
    &Q^v_\texttt{dec}\!\!W^q_\texttt{dec} \mathbf{F}^v_n,\quad \!\! K^{va}_\texttt{dec}\!=\!W^k_\texttt{dec} \mathbf{F}^{va},\quad \!\! S^{va}_\texttt{dec}\!=\!W^s_\texttt{dec} \mathbf{F}^{va} \!\! \\
    &\mathbf{\tilde{F}}^v=\mathrm{softmax}(\frac{Q^v_\dec {K^{va}_\dec}^{\mathrm{T}}}{\sqrt{d_k}})S^{va}_\dec
\end{align}
%It can be formulated as,
%\begin{align}
%    &\mathbf{\tilde{F}}^{v}=\phi_{attn}(W^{q_v} \mathbf{F}^v, W^{k_v} \mathbf{F}^{av}, W^{s_v} \mathbf{F}^{av}), \\
%    &\mathbf{\tilde{F}}^{a}=\phi_{attn}(W^{q_a} \mathbf{F}^a, W^{k_a} \mathbf{F}^{av}, W^{s_a} \mathbf{F}^{av}),
%\end{align}
%where $W^{q_v},W^{q_a},W^{k_v},W^{k_a},W^{s_v},W^{s_a}$ are learnable parameters and used to linearly transform the input to query space, key space and value space respectively.
where $W^q_\dec, W^k_\dec, W^s_\dec$ are learnable parameters and used to linearly transform the input to the query, key and value.
The other decoder is also applied to exploit the associations between the audio features  $\mathbf{F}^a_n$ and the sequence modality features $\mathbf{F}^{va}$ and then generate the co-occurrence representations $\mathbf{\tilde{F}}^a$.
\subsection{Highlight Estimation}
Next, the intra-modality representations $\mathbf{\hat{F^v}},\mathbf{\hat{F^a}}$ and the cross-modality co-occurrence representations $\mathbf{\tilde{F}}^{v}, \mathbf{\tilde{F}}^{a}$ are regarded as the latent feature vectors for highlight detection.
Since the information across different modalities provide rich highlight-matter signals, we fuse the cross-modality representations $\mathbf{\tilde{F}}=\mathrm{concat}(\mathbf{\tilde{F}}^{v}, \mathbf{\tilde{F}}^{a})$.
After that, several layers formed by 
{Linear \!\! $\rightarrow$ \!\! ReLU \!\! $\rightarrow$ \!\! Dropout \!\! $\rightarrow$ \!\! Linear}
are used to transform the embedding vectors $\mathbf{\tilde{F}}, \mathbf{\hat{F}}^v, \mathbf{\hat{F}}^a$ into highlight scores $\tilde{y}, \hat{y}^v,\hat{y}^a \in \mathbb{R}^{T} $ respectively 
Finally, a weighted sum of these scores are produced to detect highlight moments.

%TODO  consider the background????
\subsection{Hard-Pairs Guided Contrastive Learning}
\label{sec:3-7}
Despite impressive performance achieving by segment cross-entropy (CE) loss in previous methods \cite{hong:2020:mini=net,badamdorj:2021:joint}, it still suffers from two limitations:
(1) the CE loss independently optimizes segment-level predictions while not considering the relationships between segments especially in temporal video sequence.
(2) the CE loss cannot concentrate on inducing strong intra-class compactness which benefits to feature discrimination.
%(2) Predictions heavily depend on the CE loss optimization, it is essential to shape structural information in video embedding space in a discriminative manner.
Therefore, we introduce a hard-pairs guided contrastive learning (HPCL) by extending the unsupervised contrastive loss to a dense segment-level supervised scheme with hard-pairs mining strategy.
We first review the contrastive formulation in unsupervised setting and then describe our contrastive loss and hard-pairs sampling strategy in supervised setting.

\subsubsection{\textbf{Background}}
Contrastive learning is always applied to self-supervised learning and unsupervised learning \cite{he:2020:moco,chen:2020:mocov2,chen:2020:simplecontrats}, which both aim to learn good representation from unlabeled data.
It conducts training by gathering similar data and pushing away dissimilar data for feature discrimination.
%These methods view the original data as anchor and generate multiple views of each data as the positive views while the others are treated as negatives.
And the contrastive learning can be considered as dictionary look-up \cite{he:2020:moco,chen:2020:mocov2}.
Assume the query $q$ in a image, it generates different augmented positive version (key $k_+$) using the same image and the negative keys $k_-$ in another images.
Contrastive learning employs InforNCE \cite{van2018:infonce,gutmann2010noise} as follows,
\begin{equation}
    \mathcal{L}_\mathrm{nce}=-\log \frac{exp(q \cdot k_+/\tau)}{exp(q \cdot k_+ /\tau)+\sum_{k_-}{exp(q \cdot k_-/\tau)}}
\end{equation}
where $\tau$ represents temperature coefficient \cite{wu:2018un-featdis}.

\subsubsection{\textbf{Segment-wise Contrastive Loss}}
Our HPCL replaces the current image-wise training strategy with a segment-to-segment intra-video dense paradigm.
The HPCL is used to regularize the output feature embedding space of labeled data.
In training phase, given a target video sequence containing $T$ segments with labels $\{y_i\}_{i=1}^T$, we first aggregate the input embeddings $\mathbf{\hat{F}}^{v}, \mathbf{\hat{F}}^{a},\mathbf{\tilde{F}}^{v},\mathbf{\tilde{F}}^{a}$ into the segment-wise representations, which can be formulated as,
\begin{equation}
    \mathbf{\hat{F}}=\mathrm{concat}(\mathbf{\hat{F}^{v}}+\mathbf{\tilde{F}}^{v},\mathbf{\hat{F}}^{a}+\mathbf{\tilde{F}}^{a})
\end{equation}
where $\mathbf{\hat{F}}= \{\hat{f}_i\}_{i=1}^{T} \in \mathbb{R}^{T \times 2d} $ denotes the final segment-wise representations.
$\mathrm{concat}$ is a feature concatenate operation.
Then, for the segment $\textit{query}$ with label $y$, the positive $\textit{keys}$ are the other segments labeled $y$ while the negative $\textit{keys}$ are the segments belonging to the other class.
Our dense segment-level loss aims to contrast positive keys against negative ones.
Formally, it can be defined as,

\begin{equation}
    \scriptsize
    \mathcal{L}_\mathrm{HPCL}\!=\!
    \frac{1}{|T|}\!\sum_{q \in \hat{F}} \frac{1}{\Gamma^P_q}  \sum_{k_+ \in \Gamma^P_q} \!-\!\log  \frac{exp(q \cdot k_+/\tau)}{exp(q \cdot k_+)\!+\!\sum\limits_{k_- \in \Gamma^N_q} exp(q \cdot k_-/\tau)}
\end{equation}
where the video sequence contains $T$ segments, $\Gamma^P_q, \Gamma^N_q$ represent the segment representation sets of positive and negative keys for the query $q \in \mathbf{\hat{F}}$ separately.
Our contrastive loss function tailored for segment-wise predictions proposes to learn structure-aware representations and make the positive data closer while negative ones apart, which is useful for highlight detection.

\begin{table*}[t]
\begin{spacing}{1.0}
\centering
\scriptsize

\setlength{\tabcolsep}{2.1mm}
\begin{tabular}{c|ccccc|ccccc}
    \toprule[1.0pt]
    \multirow{2}{*}{Category} & \multicolumn{5}{c|}{Uni-Modality} & \multicolumn{5}{c}{Multi-Modality} \\
    \cline{2-11}
     & RRAE~\cite{RRAE}& LIM-s~\cite{LiM-s} & Video2GIF~\cite{gygli:2016:video2gif} & LSVM~\cite{LSVM} & SL\cite{xu:2021:crossSL} & MN\cite{hong:2020:mini=net} &Joint-VA \cite{badamdorj:2021:joint} & TCG \cite{ye:2021:TCG} & Ours &Ours*\\
    \hline
    dog & 0.49 & 0.579 & 0.308 & 0.60 & \textbf{0.708} & 0.582 & 0.645 & 0.553 & 0.678 & 0.690 \\
    gymnastics & 0.35 & 0.417 & 0.335 & 0.41 & 0.617 & 0.528 & \textbf{0.719} & 0.626 & 0.681 & 0.660 \\
    parkour & 0.50 & 0.670 & 0.540 & 0.61 & 0.772 & 0.702 & 0.808 & 0.709 & 0.791 & \textbf{0.890} \\
    skating & 0.25 & 0.578 & 0.554 & 0.62 & 0.725 & 0.722 & 0.620 & 0.691 & 0.740 & \textbf{0.741} \\
    skiing & 0.22 & 0.486 & 0.328 & 0.36 & 0.661 & 0.587 & \textbf{0.732} & 0.601 & 0.719 & 0.690 \\
    surfing & 0.49 & 0.651 & 0.541 & 0.61 & 0.762 & 0.651 & 0.783 & 0.598 & \textbf{0.822} & 0.811 \\
    \hline
    Average & 0.383 & 0.564 & 0.464 & 0.536 & 0.693 & 0.644 & 0.718 & 0.630 & 0.739 & \textbf{0.747} \\
    \bottomrule[1.0pt]
\end{tabular}
\vspace{0.17cm}
\caption{Experimental results comparisons of highlight detection on YouTube Highlight dataset in terms of mAP.
%\textbf{*} represents that we utilize \cite{Carreira:2017:I3D} as visual feature extractor.}\
Notice that the model `Ours' utilizes 3D CNN \cite{hara:2018:3dcnn} as visual feature extractor following previous work \cite{badamdorj:2021:joint,hong:2020:mini=net}, while `Ours*' uses I3D \cite{Carreira:2017:I3D} to extract visual features.
Uni-Modality represents the methods that only employing visual features while Multi-Modality represents those visual-audio methods.
 }
 \label{youtube_highlights}
\end{spacing}
\end{table*}

\begin{table*}[t]
\begin{spacing}{1.1}
\centering
\scriptsize
%\small

\setlength{\tabcolsep}{1.58mm}
\begin{tabular}{c|ccccccc|ccccc}
    \toprule[1.0pt]
    \multirow{2}{*}{Category} & \multicolumn{7}{c|}{Uni-Modality} & \multicolumn{5}{c}{Multi-Modality} \\
    \cline{2-13}
     & vsLSTM~\cite{vslstm} & SM~\cite{SM} & VESD~\cite{VESD} & LIM-s~\cite{LiM-s} & KVS~\cite{KVS} & DPP~\cite{DPP} &  SL \cite{xu:2021:crossSL}& MN\cite{hong:2020:mini=net} &  Joint-SA \cite{badamdorj:2021:joint} &TCG \cite{ye:2021:TCG} &Ours &Ours*\\
    \hline
    VT  & 0.411 & 0.415 & 0.447 & 0.559 & 0.353 & 0.399 & 0.865 & 0.806 & 0.837 & 0.850 & 0.894 & \textbf{0.908} \\
    VU  & 0.462 & 0.467 & 0.493 & 0.429 & 0.441 & 0.453 & 0.687 & 0.683 & 0.573 & 0.714 & 0.714 & \textbf{0.728} \\
    GA  & 0.463 & 0.469 & 0.496 & 0.612 & 0.402 & 0.457 & 0.749 & 0.782 & 0.785 & 0.819 & 0.844 & \textbf{0.846} \\
    MS  & 0.477 & 0.478 & 0.503 & 0.540 & 0.417 & 0.462 & 0.862 & 0.818 & 0.861 & 0.786 & 0.795 & \textbf{0.850} \\
    PK  & 0.448 & 0.445 & 0.478 & 0.604 & 0.382 & 0.437 & 0.790 & 0.781 & 0.801 & \textbf{0.802} & 0.779 & 0.783 \\
    PR  & 0.461 & 0.458 & 0.485 & 0.475 & 0.403 & 0.446 & 0.632 & 0.658 & 0.692 & 0.755 & 0.743 & \textbf{0.780} \\
    FM  & 0.452 & 0.451 & 0.487 & 0.432 & 0.397 & 0.442 & 0.589 & 0.578 & 0.700 & 0.716 & 0.704 & \textbf{0.728} \\
    BK  & 0.406 & 0.407 & 0.441 & 0.663 & 0.342 & 0.395 & 0.726 & 0.750 & 0.730 & \textbf{0.773} & 0.761 & 0.771 \\
    BT  & 0.471 & 0.473 & 0.492 & 0.691 & 0.419 & 0.464 & 0.789 & 0.802 & \textbf{0.974} & 0.786 & 0.891 & 0.895 \\
    DS  & 0.455 & 0.453 & 0.488 & 0.626 & 0.394 & 0.449 & 0.640 & 0.655 & 0.675 & 0.681 & 0.703 & \textbf{0.723} \\
    \hline
    Average & 0.451 & 0.461 & 0.481 & 0.563 & 0.398 & 0.733 & 0.447 & 0.732 & 0.763 & 0.768 & 0.783 & \textbf{0.801} \\
    \bottomrule[1.0pt]
\end{tabular}
\end{spacing}
\vspace{0.17cm}
\caption{Comparison of the highlight detection performances with state-of-the-arts on the TVSum test split in terms of top-5 mAP.}
\label{tvsum}
\end{table*}

\subsubsection{\textbf{Hard-Pairs Sampling Strategy.}}

%training and infering 
Previous methods \cite{kalantidis:2020:hard,suh:2019:stochastic,oh:2016:struct:embed} verify that mining negative samples are likely to be more useful and provide significant gradient information.
In our fully supervised setting, the negative data in contrastive learning are \textit{true negative} exactly.
Thus, we would ask \textit{what makes a good negative samples in supervised learning?}
The most useful negative samples are ones that the embedding currently believes to be similar to the query since the hardest points are those close to the query, and are expected to have a high propensity to have the same label.
For example, the hard-pairs samples always emerge from nearby highlight (positive) ones because the consecutive appearance frames lead to indistinguishable embeddings between the marginal highlight segments and negative samples closest to them (\textit{treat as hard-pairs}).
In order to improve the feature discriminating power in HPCL, we first sample these hard-pairs for video sequence and then utilize the ranking loss to optimize them.
Specifically, given a video sequence with $T$ segments and positive masks $\{y_i \in \{0,1\}\}_{i=1}^T$, the water-sheds formulated as the boundaries from \textit{positives vs.\ negatives } are identified and denoted as $\{c_j\}_{j=1}^W$.
Here $c_j$ is the index of video segments and $W$ represents the number of the water-sheds.
For a water-shed $c_j$, we sample indexes according to $c_j$ including $\Upsilon_1=\{c_j-k\}_{k=1}^L$ and $\Upsilon_2=\{c_j+k\}_{k=1}^L$, where it would be replaced with $c_j$ if $c_j-k<0$. $L=3$ is the region size.
The hard-pairs are represented by $\Upsilon=\{(c_j-k,c_j+k)\}_{k=1}^L$.
The loss is employed to optimize these hard-pairs, which is formulated as,
\begin{equation}
    \mathcal{L}_\mathrm{rank}=\sum_{p \in \Upsilon }max(margin-d(p),0)
\end{equation}
where $d(p)$ represents the euclidean distance between the features  indexed by the pairs $p$.
$margin$ is a hyper-parameter. We set $margin=0.7$.

The proposed HPCL scheme and the segment-wise cross-entropy loss are complementary to each other.
They can fully exploit the meaningful features for highlight detection.
%The cross-entropy loss function is formed as :
%\begin{align}
%    \mathcal{L}(\hat{y})=\sum_{i \in T} -[y_i \log(\hat{y})+(1-y_i)\log(1-\hat{y})]
%\end{align}
For the multi-modal predicted scores $\tilde{y}, \hat{y}^v,\hat{y}^a$, the weighted sum of training target are :
\begin{align}
    \mathcal{L}_\mathrm{ce}=L(\tilde{y},y)+L(\hat{y}^v,y)+L(\hat{y}^a,y)
\end{align}
where $y$ is the target lables and $L(\dot)$ denotes the CE loss.
Thus, the overall loss function is formulated as,
\begin{align}
    \mathcal{L}_\mathrm{hld}=\lambda_1 \mathcal{L}_\mathrm{ce}+\lambda_2  \mathcal{L}_\mathrm{HPCL}+\lambda_3  \mathcal{L}_\mathrm{rank}
\end{align}
where $\lambda_1,\lambda_2,\lambda_3$ denotes the hyper-parameter to balance the terms.
We set $\lambda_1=1,\lambda_2=0.3,\lambda_3=0.1.$

\begin{sloppypar}
\section{Experiments}
In this section, we conduct extensive experiments on two challenging benchmarks, \textit{i.e.} YouTube highlights[1] and TVSum[2], to demonstrate the effectiveness of the proposed method.

\subsection{Datasets and Experimental Setting}
\noindent\textbf{YouTube Highlights \cite{sun:2014:ranking}} is composed of six video categories, \textit{i.e.} dog, gymnastics, parkour, skating, skiing and surfing, and each category has approximately 100 videos. Segment-level annotations are provided to indicate whether a segment is a highlight moment. 
We follow the training-test split of previous works \cite{badamdorj:2021:joint,xu:2021:crossSL} for model training and evaluation.

\noindent\textbf{TVSum \cite{tvsum}} is a video summarization dataset containing 10 categories with 5 videos of each category and an average of minutes per video.
Since the ground truth annotations in TVSum are given as frame-level scores, we first need to aggregate frame-level scores to obtain segment-level scores, and then select the higher 50\% of the scores as the segment-level highlight annotations. 
For each category of 5 videos, following \cite{xu:2021:crossSL}, we choose the longer two videos for training and the remaining three videos for testing.

\noindent\textbf{Evaluation Metrics: } On the YouTube Highlights dataset, we adopt mean Average Precision(mAP), which is widely adopted in previous methods \cite{xu:2021:crossSL,badamdorj:2021:joint,hong:2020:mini=net}, as the evaluation metric. 
Unlike in image object detection, where mAP is obtained by accumulating and computing the average precision over all images, highlight detection compute mAP for each video separately, because highlight moments in one video are not necessarily more interesting than non-highlight moments in other videos. 
On the TVSum dataset, we refer to previous works \cite{xu:2021:crossSL,badamdorj:2021:joint,ye:2021:TCG} and compute the mAP at top-5 scores for every video.

%\subsection{Experimental Setup}
%\noindent\textbf{Feature representation:} In data preparation, we extract all the frames from raw videos and split them into preset clips. Then we use a 3D CNN with ResNet-34 backbone pretrained on Kinetics-400 dataset to extract the visual features. In addition, the two-stream I3D pre-trained on Kinetics was also used as our visual extractor and obtain better performance. And the temporal stride of all the above extractor is 16 frames. 

%For audio features, we first convert the video to audio format using ffmpeg. As in [1], we exploit the PANN audio network[22] pretrained on AudioSet to obtain audio features aligned with each visual clip from the auido files. Finally, both visual and audio features are average pooled within each segment to acquire the segment-level feature.
\end{sloppypar}

\begin{figure*}[t]
%\begin{wrapfigure}{r}{0.5\textwidth}
  \centering
  \includegraphics[width=1.0\linewidth]{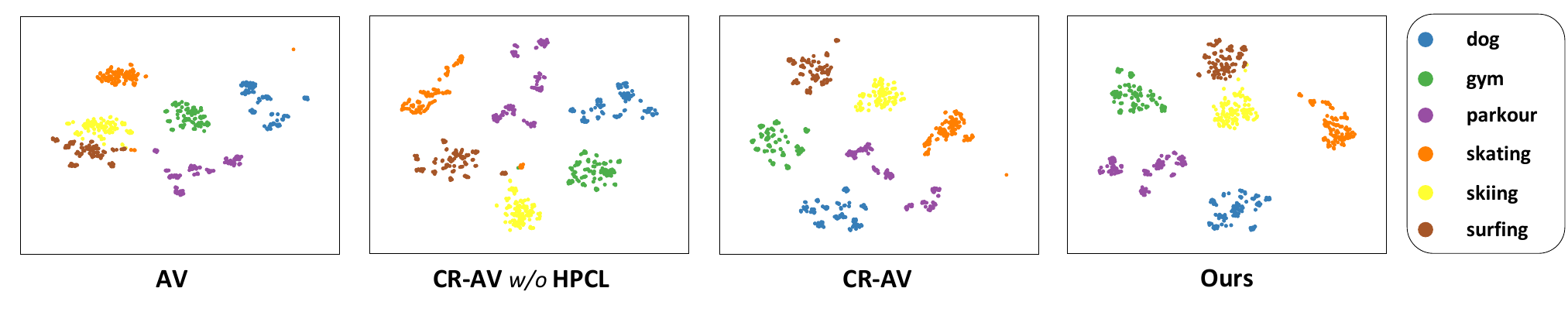}
  \caption{
  t-SNE plots of feature embedding for the testing split of YouTube Highlights.
  We visualize four architecture variants for better comparison.
  Each segment-level embedding is viewed as a point and the segment belonging the same category have the same color. 
  Best view in color.
  }
  \label{fig:visual1}
\end{figure*}

\begin{sloppypar}
\subsection{Implementation details}
%data preprcess
In data pre-process, we split the untrimmed videos into several segments following the previous works \cite{badamdorj:2021:joint,xu:2021:crossSL}.
Since the highlight detection datasets are highly imbalanced, we make sure the $T$ selected video segments in a video sequence contain both highlight (\textit{positive}) and non-highlight (\textit{negative}) segments for contrastive learning and hard-pairs ranking.
Specifically, in the sampling process, the training samples need to meet certain conditions. 
First, we sample $T=20$ segments for a video sequence where the segment sequence must contain both positive samples and negative samples, and the ratio is not less than 1:2.
Secondly, if the above is not satisfied, we re-sample additional samples in the same video, and the final sampling results must satisfy the temporal order.
In our model, we use $n=2$ transformer encoder/decoder layers with 8 attention heads blocks for intra-modality and cross-modality co-occurrence encoding and set dropout probability to $0.5$.
The embedding dimension $d=256, d_k=512,d_v=512$.
Adam optimization algorithm is applied for training the models with 20 epochs on both Youtube Highlights and TVSum benchmarks.
The learning rate is set to 1e-4. 
%We provide the details in Supplemental Material.
Finally, all our experiments are conducted on a single NVIDIA Tesla V100 GPU. 
%And our model is implemented with the Pytorch deep learning framework.
\end{sloppypar}

\begin{table}[t]
\centering
\small
\scalebox{1.0}{
\begin{tabular}{c|cc}
\hline
%Variants&Dataset&Average Results\\
\multirow{2}{*}{{Architecture Variants}} &\multicolumn{2}{c}{{Average Results}}\\
\cline{2-3}
&YouTube Highlights &TVSum \\
%\multirow{1}{|c|}{{Variants}} &\multirow{2}{c|}{{Dataset}}& \multicolumn{2}{c}{Average Results} \\
%1 && Baseline &  & 41.76 & 32.17 & 21.72\\
\hline
{V Only} &0.659    &0.763  \\
{A Only} &0.651    &0.752  \\
{AV}     &0.675      &0.784  \\
{CR-AV}  &0.697 &0.789 \\
{CO-AV~(ours)} &\textbf{0.747} &\textbf{0.801}\\
\bottomrule
\end{tabular}
}
\caption{Ablation Study on the various modifications of our proposed method.}
\label{tab:viariants}
\end{table}

\begin{table}[t]
\centering
\small
\scalebox{1.0}{
\begin{tabular}{c|cc}
\hline
%Variants&Dataset&Average Results\\
\multirow{2}{*}{{Learning Scheme}} &\multicolumn{2}{c}{{Average Results}}\\
\cline{2-3}
&YouTube Highlights &TVSum \\
%\multirow{1}{|c|}{{Variants}} &\multirow{2}{c|}{{Dataset}}& \multicolumn{2}{c}{Average Results} \\
%1 && Baseline &  & 41.76 & 32.17 & 21.72\\
\hline
$\mathcal{L}_\mathrm{ce}$ (baseline) &0.702    &0.766  \\
$\mathcal{L}_\mathrm{ce}$+$\mathcal{L}_\mathrm{HPCL}$  &0.733  &0.792  \\
$\mathcal{L}_\mathrm{ce}$+$\mathcal{L}_\mathrm{HPCL}$+$\mathcal{L}_\mathrm{rank}$  &\textbf{0.747}&\textbf{0.801}  \\
\bottomrule
\end{tabular}
}
\caption{Ablation Study on the effect of dense contrastive learning scheme.}
\label{tab:loss}
\end{table}

\subsection{Comparison with the State of the Art}
We compare our proposed method with the other state-of-the-art methods \cite{RRAE,LiM-s,gygli:2016:video2gif,LSVM,badamdorj:2021:joint,ye:2021:TCG,hong:2020:mini=net,xu:2021:crossSL,vslstm} on two widely adopted benchmarks, \ie, YouTube Highlights and TVSum.
%%need check TODO
Specifically, we also present the results using visual feature extractor I3D\cite{Carreira:2017:I3D}.
\begin{sloppypar}

\noindent\textbf{Results on YouTube Highlights.} 
The results are listed in Table \ref{youtube_highlights}.
Our methods achieve superior performance compared with all of the aforementioned methods with a considerable margin when we adopt the same visual feature extractor 3DCNN\cite{hara:2018:3dcnn} as the works \cite{badamdorj:2021:joint,hong:2020:mini=net}.
For instance, our method improves the mAP of \textit{skating} from 0.620 in Joint-VA to 0.740.
The performance of \textit{parkour} is boosted from 0.808 to 0.890.
It indicates that fully exploiting intra-modality and inter-modality relations benefit the detection result.
%Our method with unique visual signals also outperforms many previous uni-modality works.
The average result can be further improved by 0.8\% when we use the I3D \cite{Carreira:2017:I3D} backbone for visual features.
%Compared with the other methods, our method perform better in almost all categories.
%It can be observed that our method outperforms both ranking-based[1,2, 3, 4] and set-based[5, 6, 7] methods proposed in previous works.
%And in three of all categories in YouTube Highlights dataset, our method perform best.
%Our proposed method significantly outperforms all previous highlight detection methods in three categories of the YouTube Highlights dataset. 
%Especially in the two categories of parkour and skating, which exceeds all other methods by a lot. 
%Overall, our method achieves the highest average mAP across all video categories, outperforming all previous methods.

\begin{figure*}[t]
%\begin{wrapfigure}{r}{0.5\textwidth}
  \centering
  \includegraphics[width=1.0\linewidth]{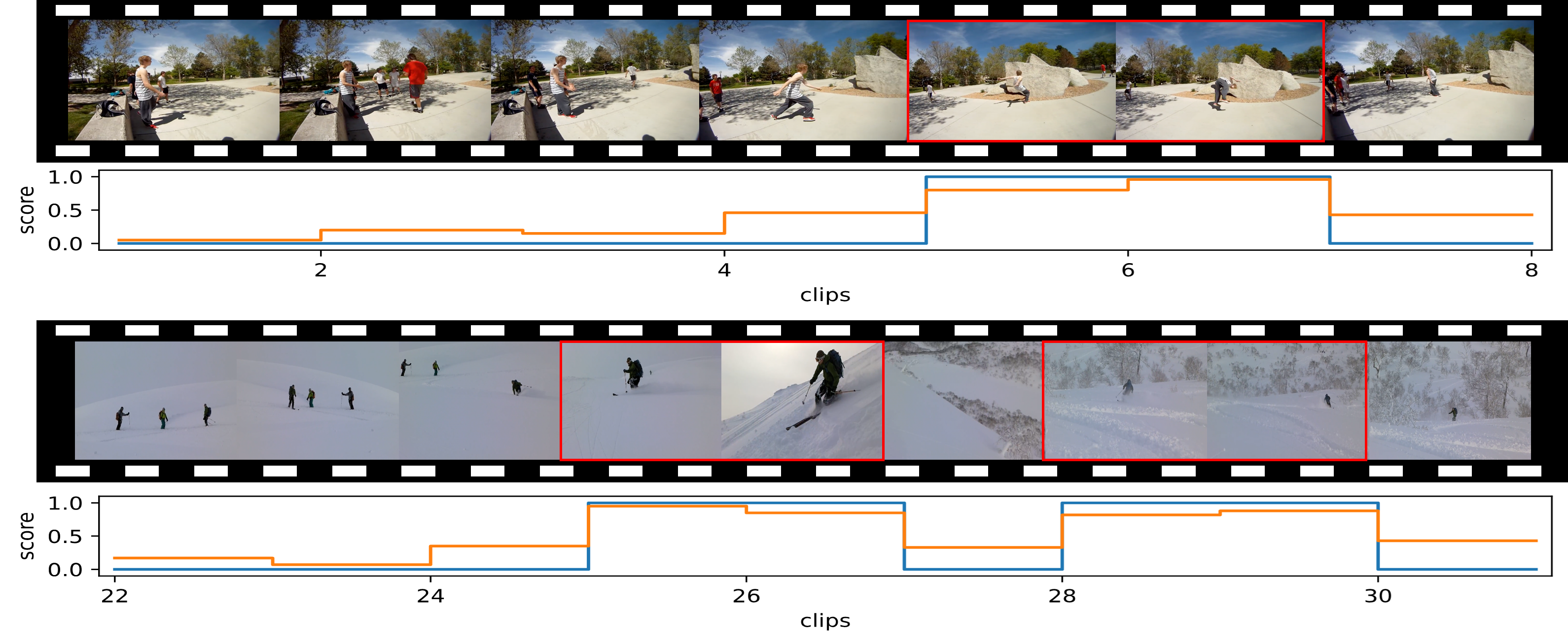}
  \caption{Qualitative results.
  We show highlight detection results on the test set of YouTube Highlights. The red box represent the ground truth segments.
  }
  \label{fig:visual2}
\end{figure*}

\noindent\textbf{Results on TVSum.}
We also provide the detailed comparisons with previous works as shown in Table \ref{tvsum}.
The results with visual features extracted by \cite{hara:2018:3dcnn} can reach 0.783, which outperforms most methods with the same backbones \cite{badamdorj:2021:joint,hong:2020:mini=net}.
A considerable improvement is achieved by using the backbone I3D \cite{Carreira:2017:I3D}.
We speculate that the I3D captures high-level features with larger receptive field, which benefits our feature discrimination using HPCL.
It is noting that the corss-attention method \cite{badamdorj:2021:joint} achieves $0.763$ and performs lower than our performance $0.801$, indicating the benefits of intra-modality and inter-modality learning and feature discrimination over the simple cross-attention mechanism.
\end{sloppypar}

\subsection{Ablation Studies}
In this section, we conduct ablation studies on both two datasets to verify the effectiveness of the main components.
The average results for all categories in a dataset are displayed for comparisons.

\begin{table}[t]
\centering
\small
\scalebox{0.9}{
\begin{tabular}{c|cc|cc}
\hline
%Variants&Dataset&Average Results\\
\multirow{2}{*}{{Methods}} &\multicolumn{2}{c}{{Initial Manner}}&\multicolumn{2}{|c}{{Average Results}}\\
\cline{2-5}
& \textit{Random} & \textit{Mean} & Youtube Highlight & TVSum \\
%\multirow{1}{|c|}{{Variants}} &\multirow{2}{c|}{{Dataset}}& \multicolumn{2}{c}{Average Results} \\
%1 && Baseline &  & 41.76 & 32.17 & 21.72\\
\hline
Ours \textit{w/o} $\mathbf{G}$ &  &  &0.710 &0.789 \\
Ours \textit{w/} $\mathbf{G}$ & \cmark  &  &0.744 &\textbf{0.803} \\
Ours \textit{w/} $\mathbf{G}$ & & \cmark &\textbf{0.747} &0.801 \\

\bottomrule
\end{tabular}
}
\caption{Ablation Study on the effect of Global Representation.
\textit{random} represents randomly initialize global embedding while \textit{Mean} utilizes the mean of input segment-wise embedding.}
\label{tab:global}
\end{table}

\subsubsection{Architectures Variants}
\label{abl:arch:variants}
\begin{sloppypar}
To intuitively show the effectiveness of the proposed method, we present the following various modifications of our proposed methods:
1)~\textbf{A (V) Only:} we only utilize the audio (visual) signals for feature learning in our work and discard the visual (audio) stream and co-occurrence encoding.
2)~\textbf{AV:} the visual and audio features extracted by feature extrator are simply aggregated by concatenation and then projected into the intra-modality module for feature modeling.
3)~\textbf{CR-AV:} following the implementation of \cite{badamdorj:2021:joint}, we employ the cross-attention blocks to our cross-modality module for the audio-visual signals modeling.
4)~\textbf{CO-AV}: Our final architecture with intra-modality and cross-modality co-occurrence encoding.
It is worth noting that all variants introduce the HPCL for model optimization.
The model setting follows our final architectures for all modifications.
Table \ref{tab:viariants} summarizes the results of these architectures variants.
The cross-modality representation modeling can generally improve the performance from 0.675 to 0.697 in YouTube Highlights as shown in the third and forth rows in Table \ref{tab:viariants}.
Furthermore, compared to the \textbf{CR-AV}, our method with co-occurrence encoding (\textbf{CO-AV}) can boost the performance from 0.697\% to 0.747\% in YouTube Highlights dataset, showing the superiority and effectiveness of our intra-modality and cross-modality co-occurrence representation encoding.
\end{sloppypar}

\vspace{-0.3cm}
\subsubsection{Effect of HPCL}
We validate the design of our HPCL scheme as shown in Table \ref{tab:loss}.
We formulate that the \textbf{Baseline} discards the contrastive loss and hard-pairs rank loss and only utilizes segments-wise cross-entropy loss for highlight detection.
The results between the first and second rows suggest that applying contrative loss in supervised setting improve the performance from 0.702 to 0.733 in YouTube Highlights and 0.766 to 0.792 in TVsum, which demonstrate the superiority of our HPCL.
Also, the hard-pairs sampling is further boost the performance by 1.4\%, validating our analysis that mining hard pairs is helpful for discriminating power improvement.

\subsubsection{Effect of Global Representation}
In this experiment, we verify that global representation play an important role for semantic feature exploiting as shown in Table \ref{tab:global}.
As we can see, there is a little difference between the performances of the model with or without global decoder. 
The model using random initial global embedding performs slight worse than using the mean of segment-wise features.
%there is little to no difference between different ways of encoding positional information.

\subsection{Visualization}
\textbf{Feature Distribution Visualization.}
We apply the t-SNE \cite{van2008:tsne} to the aggregated visual and audio representations on the YouTube Highlights dataset.
Figure \ref{fig:visual1} displays the t-SNE visualization of our architecture variants as illustrated in Sec. \ref{abl:arch:variants}.
We find \textbf{CR-AV} \textit{w/ HPCL} performs well compared to the original \textbf{CR-AV} \textit{w/o} HPCL, showing the strong intra-class compactness and inter-class dispersion .
In addition, when integrating the HPCL and cross-modality co-occurrence as our final model, the features are better separated.

\noindent
\textbf{Qualitative Results.}
As shown in Figure \ref{fig:visual2}, we display some qualitative results on the YouTube Highlights dataset.
Our proposed method can successfully detect the shining moments, and the highlight moments and background scenes can be well distinguished.

\section{Conclusion}
This paper proposes a novel highlight detection methods, which aims to pursue two confounding goals: 1) cross-modal representation learning; 2) inter-segments feature discrimination.
In the former case, we propose a visual-audio network to explore cross-modal representation by measuring the relations from intra-modality and inter-modality with the modal-wise and sequence-wise attention mechanism.
To enhance the video representation, we also introduce a global decoder to abstract global informative features by selectively integrating the segment-level representations.
In the later case, a hard-pairs guided contrastive learning scheme is introduced to shape segment representations by improving intra-class compactness and inter-class dispersion in a discrminative manner with hard-pairs sampling strategy.
%We also present a hard-pairs sampling strategy to improve the discriminating power by mining confused hard-pairs.
Extensive experiments conducted on two widely adopted benchmarks demonstrate the effectiveness and superiority of our proposed method compared to previous methods.
{\small
\bibliographystyle{ieee_fullname}
\bibliography{article}
}

\end{document}